\documentclass[11pt,a4paper]{article}
\usepackage[hyperref]{acl2021}
\usepackage{times}
\usepackage{latexsym}

\usepackage{microtype}

\aclfinalcopy 

\usepackage{soul}      
\usepackage{colortbl}  
\usepackage{amsmath}   
\usepackage{amsfonts}  
\usepackage{amssymb}   
\usepackage{enumitem}
\usepackage{graphicx}
\usepackage{multirow}
\usepackage{colortbl}
\usepackage{rotating, graphicx}
\usepackage{multirow, makecell, caption}
\usepackage{booktabs}
\usepackage{url}

\usepackage{expl3, xparse} 
\ExplSyntaxOn
\cs_new:Npn \__my_use:N #1
  {
    \quark_if_recursion_tail_stop:n {#1}
    \cs_if_exist_use:cF { l__my_#1_fp } {#1}
    \__my_use:N
  }
\cs_new:Npn \my_use:n #1
  { \fp_eval:n { \__my_use:N #1 \q_recursion_tail \q_recursion_stop } }
\cs_new_protected:Npn \my_set:nn #1#2
  {
    \fp_zero_new:c { l__my_#1_fp }
    \fp_set:cn { l__my_#1_fp } { \my_use:n {#2} }
  }

\ExplSyntaxOff

\title{Multilingual Answer Sentence Reranking via\\ Automatically Translated Data}

\author{Thuy Vu \\
  Amazon Alexa AI\\ Manhattan Beach, California, USA \\
  \texttt{thuyvu@amazon.com} \\\And
  Alessandro Moschitti \\
  Amazon Alexa AI\\ Manhattan Beach, California, USA \\
  \texttt{amosch@amazon.com} \\}

\date{}

\definecolor{mygreen}{rgb}{0.0, 0.44, 0.0}

\newcommand{\GPD}{{$D$}}

\newcommand{\ASS}{AS2}

\begin{document}
\maketitle
\begin{abstract}
We present a study on the design of multilingual Answer Sentence Selection ({\ASS}) models, which are a core component of modern Question Answering (QA) systems. The main idea is to transfer data, created from one resource rich language, e.g., English, to other languages, less rich in terms of resources.
The main findings of this paper are: (i) the training data for AS2 translated into a target language can be used to effectively fine-tune a Transformer-based model for that language; (ii) one multilingual Transformer model it is enough to  rank answers in multiple languages; and (iii) mixed-language question/answer pairs can be used to fine-tune models to select answers from any language, where the input question is just in one language. This highly reduces the complexity and technical requirement of a multilingual QA system.
Our experiments validate the findings above, showing a modest drop, at most 3\%, with respect to the state-of-the-art English model.

\end{abstract}

\section{Introduction}

Document retrieval-based QA is essentially based on (i) a search engine, which retrieves documents relevant to a question, and (ii) an {\ASS} component, which selects the most probable sentence candidate as the answer. This approach can be improved using Machine Reading models \cite{chen2017reading,das2019building}, which also poses some efficiency challenges to process hundreds of document candidates. 
{\ASS} has been continuously improved for English~\cite{garg2019tanda}, where such results have been mainly driven by the availability of large and curated training data, besides other technical innovations on the neural network front.
One drawback of the current data-driven approach is that the effort on one language, carried out to create training data, has to be replicated for other languages. One simple idea for building multi-language QA systems is the use of automatically translated training data.

In this paper, we study the possibility of using automatic Machine Translation (MT) to transfer data from one language to another for the design of multilingual QA.
Specifically, we create a large scale QA dataset for English of $\sim$120K question-answer pairs, named QAD and denoted as $\text{\GPD}^{\text{En}}$.
Then, We transfer $\text{\GPD}^{\text{En}}$ into different languages to build multilingual {\ASS} models for selecting answer sentences.
We show that on an average, Transformer-based models \cite{devlin2018bert} fine-tuned on training data translated into the target language drop $\sim$10\% of their Accuracy. This happens both for monolingual and cross-lingual Transformer models.

We thus propose and test two strategies to improve the previous methods:
(i) fine-tuning with the translated data in more languages, instead of just on the target one, which produces an average gain of $\sim$3\%;
(ii) to combine question-answer pairs with pairs formed by members of different languages, which reduces the accuracy gap to a state-of-the-art monolingual performance on English, by just 2\% (i.e., other 5\% of gain).
The interesting aspect of the latter finding is that, not only we can ask questions in different languages, but we can also answer in different languages, just using the question in one language. That is, using one Transformer-based model, we can ask a question in one language and select answer candidates in sources from different languages. Then, it is always possible to translate the answer in the enquired language: this is a remarkable advantage over multi-language systems based on MT, which require both to translate the question and the answer.

\section{Answer Sentence Selection}
\label{sec:problem}
Cross-language retrieval is an important component of a multilingual QA. However, we can assume from previous work \cite{DBLP:journals/corr/abs-1710-01487} to have a search engine retrieving documents in different languages. Thus, we focus on the most innovative aspects of multilingual QA, which is {\ASS} based on pre-trained Transformer models.
\subsection{Definition}
\label{sec:definition}
The task of reranking answer sentence candidates provided by a retrieval engine can be modeled with a classifier scoring answer candidates for a question.
Let $q$ be a question, $T_q=\{t_1, \dots, t_n\}$ be a set of answer candidates for $q$, we define $\mathcal{R}$ a ranking function that orders candidates in $T_q$ according to a score $p\left(q, t_i\right)$, indicating the probability of $t_i$ to be a correct answer for $q$.
Popular methods modeling $\mathcal{R}$ include Compare-Aggregate~\cite{DBLP:journals/corr/abs-1905-12897}, inter-weighted alignment networks~\cite{shen-etal-2017-inter}, and BERT~\cite{garg2019tanda}.

\subsection{Transformer Model for {\ASS}}
\label{sec:model}

Transformer-based architectures have proved to be powerful language models, which can capture complex linguistic regularities and semantic similarity  patterns. Thus, they benefit a wide range of NLP tasks, including {\ASS}.

Let $\mathcal{B}$ be a pre-trained language model, e.g., the recently proposed BERT~\cite{devlin2018bert}, RoBERTa~\cite{DBLP:journals/corr/abs-1907-11692}, XLNet~\cite{DBLP:journals/corr/abs-1906-08237}, AlBERT~\cite{lan2019albert}.
We use $\mathcal{B}$ to compute the embedding representation of the tuple members: $\mathcal{B}\left(q, t\right) \rightarrow {\bf x} \in \mathbb{R}^d$,
where $\left(q, t\right)$ is a (question, candidate) tuple, ${\bf x}$ is the output representation of the  pair, and $d$ is the dimension of the output representations.
The classification layer is a standard feedforward network as $\mathcal{N}\!\left({\bf x}\right) = {\bf W}^{\intercal}{\bf x} + b$, where {\bf W} and $b$
are parameters we learn by fine-tuning the model on a dataset $D$.

\section{Data Modeling}
\label{sec:data}

We build a large scale annotated dataset for {\ASS}, named Question Answering Datasets (QAD).
The dataset contains a set of anonymized information inquiry questions, sampled from Alexa Traffic.
Each question has up to 100 candidates extracted from relevant web documents.
We first describe the construction of QAD, and then explain how we transfer it from one language to another.

\subsection{Question Answering Datasets}
\label{sec:qadata}

Given a question, we use a search engine with a large index to retrieve relevant documents.
Specifically, we retrieved high-probably relevant candidates as follows: we (i) retrieved top 500 relevant documents; (ii) automatically extracted the top 100 sentences ranked by a BERT model over all sentences of the documents; and (iii) had them manually annotated as correct or incorrect answers.
This process does not guarantee that we have all possible correct answers from the documents but the probability to miss them is much lower than for other datasets, only using a simple search engine for sentence retrieval. This dataset is richer than standard {\ASS} datasets, e.g., WikiQA \cite{yang-etal-2015-wikiqa}, as it consists of answers from multiple sources in addition to Wikipedia. Furthermore, the average number of answers to a question is also higher than other {\ASS} datasets. Table~\ref{data:as2} shows the statistics of the dataset.

\begin{table}
\centering
\resizebox{0.8\linewidth}{!}{%
\begin{tabular}{|c|l|r|r|r|} 
\hline
\multicolumn{1}{|l|}{} & data split & \multicolumn{1}{l|}{\#Qs} & \multicolumn{1}{l|}{\#As} & \multicolumn{1}{l|}{\#wrong-As}  \\ 
\hline
\multirow{3}{*}{\GPD}   & train  & 913                       & 24,558                     & 69,142                           \\ 
\cline{2-5}
                        & dev    & 551                       & 4,391                     & 9,509                           \\ 
\cline{2-5}
                        & test   & 427                       & 3,334                     & 7,341                           \\
\hline
\end{tabular}
}
\caption{{\GPD} Statistics}
\label{data:as2}
\vspace{-1em}
\end{table}

\subsection{Transferring QAD to other languages}
\label{sec:tqadata}

We denote our QAD dataset for {\ASS} as $D$ or in language $\ell_0$ as $D^{\ell_0}$.
$D^{\ell_0}$ can be transferred to another language $\ell_1$ by an operation $M_{\ell_0 \rightarrow \ell_1}\left(D^{\ell_0}\right): D^{\ell_0} \rightarrow D^{\ell_1}$, which in our case is a MT processing.
In addition, we can continue to transfer $D^{\ell_1}$ into $\ell_2$ to create ${D^{\ell_1}}^{\ell_2}$, with a second MT pass.
Besides, $D$ is comprised of tuples, $\left(q, t, l\right)$, where $q$, $t$, and $l$ are question, answer candidate, and label, respectively, as described in Section~\ref{sec:model}.
Interestingly, $D$ can be in mixed languages, i.e., question and answer can be in different languages. 
We denote $D^{\ell_q\ell_t}$ a dataset where questions and answers are in language $\ell_q$ and $\ell_t$, respectively.
Finally, we also denote $D^{\ell_a+\ell_b}$ the concatenation of two datasets $D^{\ell_a}$ and $D^{\ell_b}$.
We use Amazon Translate~\footnote{\url{aws.amazon.com/translate}} to translate our $\text{\GPD}^{\text{En}}$ into other languages.

\section{Experiments}
\label{sec:experiment}

We measure the performance of {\ASS} models fine-tuned using original and the transferred data described in Section~\ref{sec:data}.
We consider English (En) the source language given our label data, $\text{\GPD}^{\text{En}}=\text{\GPD}$.
We additionally consider four target languages: French (Fr), German (De), Italian (It), and Spanish (Es). All the data used in our experiments are originated from $\text{\GPD}^{\text{En}}$.
Without loss of generality, we choose German as our main target language in our experiments.
Specifically, we create {\ASS} models for English and German and study their performance gap in the following scenarios:\\
-- Single-language modeling: a separate {\ASS} model for each language, the pre-trained model can either be monolingual, e.g., specific to English or German, or multilingual.\\
-- Multi-language modeling: a shared {\ASS} model for multiple languages, using $\text{\GPD}^{\text{En}}$, and its transferred data in other languages, e.g., $\text{\GPD}^{\text{De}}$, $\text{\GPD}^{\text{It}}$, etc.\\
-- Mixed-language modeling: training on data having $q$ and $t$ in different languages, e.g., $\text{\GPD}^{\text{EnDe}}$, i.e., $q$ is in English, $t$ is in German.

\begin{table}[t]
\centering
\resizebox{0.9\linewidth}{!}{%
\begin{tabular}{|c|l|c|} 
\hline
Language                                   & \multicolumn{1}{c|}{Pre-trained Model (PT)} & Name  \\ 
\hline
\multirow{1}{*}{English}                   & bert-base-uncased                      & $\mathcal{B}_{\text{En}}$   \\ 
\hline
\multirow{1}{*}{German}                    & bert-base-german-dbmdz-cased        & $\mathcal{B}_{\text{De}}$   \\ 
\hline
\multirow{1}{*}{Multilingual} & bert-base-multilingual-cased         & $\mathcal{B}_{\text{Mu}}$   \\ 
\hline
\end{tabular}}
\caption{Pre-trained models used in our experiments}
\label{table:pretrains}
\vspace{-1em}
\end{table}

\subsection{Experiment Setting}

We conducted the experiments using the HuggingFace's Transformer library~\cite{Wolf2019HuggingFacesTS},
 using the default hyper-parameter setting of GLUE trainings: (i) AdamW variant~\cite{DBLP:journals/corr/abs-1711-05101} as optimizer, (ii) a learning rate of ~$2e\text{-}05$ set for all fine-tuning exercises, and (iii) a maximum sequence length set to 128. Our number of iterations is three.
We also use a development set to enable early stopping based on mean average precision (MAP) measure after the first iteration.
We fix the same batch size setting in the experiments to avoid possible performance discrepancies.
Random seed is set 42 by default.
Table~\ref{table:pretrains} shows all the pre-trained (PT) models we used in the experiments.

We use precision-at-1 (P@1, or accuracy), MAP, and mean reciprocal rank (MRR) as evaluation metrics.
Each experiment is characterized by: (i) PT: a pre-trained model; (ii) FT: a fine-tuning dataset with its development set: this can be $\text{\GPD}^{\text{En}}$ or its transferred data or the combination; and (iii) a test set of questions in a specific language.

\subsection{Upper-bound Performance on $\text{\GPD}^{\text{En}}$}

We first analyze the performance of the upper bound models, i.e., the state-of-the-art BERT-Base pre-trained models for English ($\mathcal{B}_{\text{En}}$) and Multilingual ($\mathcal{B}_{\text{Mu}}$), fine-tuned on original data $\text{\GPD}^{\text{En}}$.
We then compare the models using translated data, $\text{\GPD}^{\text{De}}$, and back-back-translated data, denoted  with ${\text{\GPD}^{\text{De}}}^{\text{En}}$.
We test models on either $\text{\GPD}^{\text{En}}$ or ${\text{\GPD}^{\text{De}}}^{\text{En}}$. 

\begin{table}
\centering
\resizebox{\linewidth}{!}{%
\begin{tabular}{|l|l|l|r|r|r|r|} 
\hline
\multicolumn{1}{|c|}{PT} & \multicolumn{1}{c|}{FT on {\GPD$^x$}}          & \multicolumn{1}{c|}{Test on {\GPD$^x$}}              & \multicolumn{1}{c|}{P@1} & \multicolumn{1}{c|}{MAP} & \multicolumn{1}{c|}{MRR}  \\ 
\hline
\hline
\multirow{3}{*}{$\mathcal{B}_{\text{En}}$} & {${\text{En}}$}                               &  {${\text{En}}$}   & 0\% & 0\% & 0\%  \\ 
\cline{2-6}
                      & \multirow{2}{*}{${{\text{De}}}^{\text{En}}$} &  {${\text{En}}$} & -2.3\% & -2.1\% & -2.2\%  \\ 
\cline{3-6}
                      &                                            & ${{\text{De}}}^{\text{En}}$ & -5.6\% & -4.6\% & -4.2\%  \\
\hline
\hline  
\multirow{3}{*}{$\mathcal{B}_{\text{Mu}}$} & {${\text{En}}$}                               & {${\text{En}}$}  & 0\% & -1.5\% & -0.8\%  \\ 
\cline{2-6}
                      & \multirow{2}{*}{${{\text{De}}}^{\text{En}}$} &  {${\text{En}}$} & -1.9\% & -2.3\% & -1.8\%  \\ 
\cline{3-6}
                      &                                            & ${{\text{De}}}^{\text{En}}$ & -4.7\% & -4.1\% & -3.9\%  \\ 
\hline
\end{tabular}}
\caption{Experimental results on models fine-tuned and tested using $\text{\GPD}^{\text{En}}$, or fine-tuned and tested on its back-back-translated version ${\text{\GPD}^{\text{De}}}^{\text{En}}$.}
\label{exp:naturalornot}
\end{table}

The results of Table~\ref{exp:naturalornot} shows the expected drop in performance, when moving from $\text{\GPD}^{\text{En}}$ to automatic transferred data.
Interestingly, on the natural $\text{\GPD}^{\text{En}}$ test set, the performance gap between the models trained with the original data and translated data is less than $\sim$2.3\% in P@1
for the monolingual and multilingual BERT, respectively. 
In contrast, there is an average drop of $\sim$5.2\% in P@1, when a model is both tuned and tested on translated data ${\text{\GPD}^{\text{De}}}^{\text{En}}$, confirming the importance of testing on a non noisy test set.

\subsection{Single-Language {\ASS} Modeling}

We evaluate the performance of models fine-tuned only on English using $\text{\GPD}^{\text{En}}$ or German using $\text{\GPD}^{\text{De}}$.
The models are also tested on the respective test sets.
We experimented with multiple pre-trained models for German and multilingual and selected the best one for $\mathcal{B}_{\text{De}}$ and$\mathcal{B}_{\text{Mu}}$.
The results, presented in Table~\ref{exp:singlelanguageresult}, shows a consistent performance drop on German, regardless the fact that the pre-trained model is German specific (monolingual) or multilingual.
In particular, we find that models both tuned and tested on translated data for German show an average drop of $\sim$11\% in P@1, compared to English fine-tuning. This also shows that the use of specialized pre-trained models for German does not provide significant benefits.

\begin{table}[t]
\centering
\resizebox{.9\linewidth}{!}{%
\begin{tabular}{|l|l|c|r|r|r|} 
\hline
\multicolumn{1}{|c|}{\multirow{1}{*}{PT}} & \multicolumn{1}{c|}{\multirow{1}{*}{\shortstack{FT on {\GPD$^x$}}}} & \multicolumn{1}{c|}{\multirow{1}{*}{\shortstack{Test on {\GPD$^x$}}}} & \multicolumn{1}{c|}{P@1} & \multicolumn{1}{c|}{MAP} & \multicolumn{1}{c|}{MRR}  \\ 
\hline
\hline
$\mathcal{B}_{\text{De}}$                                                & ${\text{De}}$ & ${\text{De}}$                        & -10\%                    & -6.8\%                    & -7\%                     \\ 
\hline
\hline
\multirow{2}{*}{$\mathcal{B}_{\text{Mu}}$}                               & {${\text{En}}$}                                                  & ${\text{En}}$                        & 0\%                    & -1.5\%                    & -0.8\%                     \\ 
\cline{2-6}
                                                    & ${\text{De}}$ & ${\text{De}}$                        & -11.7\%                    & -7.3\%                    & -8.8\%                     \\ 
\hline
\end{tabular}}
\caption{Result for single-language {\ASS} model}
\label{exp:singlelanguageresult}
\vspace{-1em}
\end{table}

\subsection{Multi-Language {\ASS} Modeling}

We evaluate the performance impact when fine-tuning on $\text{\GPD}^{\text{En}}$ using $\mathcal{B}_{\text{Mu}}$ together with other transferred datasets from a single pre-trained model.
Specifically, we study the performance of English and German on $\text{\GPD}^{\text{En}}$ and $\text{\GPD}^{\text{De}}$ as we add more transferred data, including $\text{\GPD}^{\text{Fr}}$, $\text{\GPD}^{\text{Es}}$, and $\text{\GPD}^{\text{It}}$ into the fine-tuning dataset.

\begin{table}
\centering
\resizebox{1\linewidth}{!}{%

\begin{tabular}{|p{5.5em}|r|r|r|r|r|r|}  
\hline
\multicolumn{1}{|c|}{\multirow{2}{*}{FT on {\GPD$^x$}}} & \multicolumn{3}{c|}{Tested on $\text{\GPD}^{\text{En}}$}                                                   & \multicolumn{3}{c|}{Tested on $\text{\GPD}^{\text{De}}$}                                                     \\ 
\cline{2-7}
\multicolumn{1}{|c|}{}                    & \multicolumn{1}{c|}{P@1} & \multicolumn{1}{c|}{MAP} & \multicolumn{1}{c|}{MRR} & \multicolumn{1}{c|}{P@1} & \multicolumn{1}{c|}{MAP} & \multicolumn{1}{c|}{MRR}  \\ 
\hline
\hline
${\text{En+De}}$ & -2.6\%                    & -3.8\%                    & -3.2\%                    & -8.9\%                    & -6.9\%                    & -7.3\%                     \\ 
\hline
${\text{En+De+Fr}}$ & -2.8\%                    & -2.8\%                    & -3.2\%                    & -7.2\%                    & -5.3\%                    & -5.5\%                     \\ 
\hline
${\text{En+De+Fr+Es}}$ & -3\%                    & -3.4\%                    & -3.3\%                    & -10\%                    & -7.4\%                 & -7.7\%                     \\ 
\hline
En+De+Fr+ Es+It & -3.7\%                    & -4.3\%                    & -3.9\%                    & -9.6\%                    & -7.1\%                    & -8.1\%                     \\
\hline
\end{tabular}}
\caption{Multi-language {\ASS} Performance using $\mathcal{B}_{\text{Mu}}$}
\label{exp:multilingual}
\vspace{-1em}
\end{table}

Table~\ref{exp:multilingual} shows that the results slightly improve.
Specifically, (i) the models both tuned on multiple languages and tested on translated data for German show smaller average drop in P@1, i.e.,  9\%, compared to the state of the art in English. This suggests that fine-tuning on multiple languages may help the performance of individual languages.
(ii) For English, there is an average drop of 3.0\%, when using $\mathcal{B}_{\text{Mu}}$, suggesting that adding language capability to the multilingual Transformer model, does not critically affect its accuracy.
This enables the use of a single multilanguage {\ASS}  model for building multilanguage QA.

\subsection{Mixed-Language {\ASS} Modeling}

Finally, we study mix language modeling, using question and candidate in different languages, both for training and testing.
We introduce two other datasets transformed from $\text{\GPD}^{\text{En}}$ and $\text{\GPD}^{\text{De}}$, namely $\text{\GPD}^{\text{EnDe}}$ and $\text{\GPD}^{\text{DeEn}}$, which contains English question and German answer pairs, as well as their swap members. 
Similarly to the previous experiment, we use $\mathcal{B}_{\text{Mu}}$ (pre-trained multilingual BERT-Base).
The results in Table~\ref{exp:mixedlanguage} show that: (i) using the mixed-language dataset for fine-tuning producs a 6\% performance drop, and 1.2\% gain, when we provide English questions, and German questions, respectively, i.e., $\text{\GPD}^{\text{En+De}}$ vs $\text{\GPD}^{\text{EnDe+DeEn}}$.
(ii) The models fine-tuned on both same-language and mixed-language pairs show stable performance. Even though we should consider the fact that the test set contains both same-language and mixed-language candidates, the performance clearly shows an improvement in dealing with German questions.
(iii) There are drops of only 
2.3\% 
and 2.8\%, 
with respect to the state-of-the-art model (original English data in Table~\ref{exp:naturalornot}), when fine-tuning on En+EnDe+ De+DeEn, for English and German test sets, respectively.

This analysis is rather interesting as enables the use of a language mix model that can be particularly useful in a production environment.
For example, multilingual QA systems based on MT, translate the question in English, select the answer, and then translate it  back to the original language.
The mix model we propose allows for saving the translation of the question. 
Most importantly, with the question in one language, we can select the candidates from all other languages.

\begin{table}[t]
\centering
\resizebox{1\linewidth}{!}{%
\begin{tabular}{|p{5em}|r|r|r|r|r|r|} 
\hline
FT & \multicolumn{1}{c|}{P@1} & \multicolumn{1}{c|}{MAP} & \multicolumn{1}{c|}{MRR} & \multicolumn{1}{c|}{P@1} & \multicolumn{1}{c|}{MAP} & \multicolumn{1}{c|}{MRR}  \\ 
\hline
\hline
\multicolumn{1}{|c|}{}                    & \multicolumn{3}{c|}{Tested on ${\text{En}}$}                                                   & \multicolumn{3}{c|}{Tested on ${\text{De}}$}                                                     \\ 
\hline
${\text{En+De}}$ & -2.6\%                    & -3.8\%                    & -3.2\%                    & -8.9\%                    & -6.9\%                    & -7.3\%                     \\ 
\hline
\hline
                                          & \multicolumn{3}{c|}{Tested on ${\text{EnDe}}$}                                                   & \multicolumn{3}{c|}{Tested on ${\text{DeEn}}$}                                                     \\ 
\hline
${\text{EnDe+DeEn}}$ & -10.5\%                    & -8.1\%                    & -9.3\%                    & -7.7\%                    & -5.6\%                    & -6.3\%                     \\ 
\hline
\hline
                                          & \multicolumn{3}{c|}{\shortstack{Tested on \\$\text{\GPD}^{\text{EnEn}}+ \text{\GPD}^{\text{EnDe}}$}}                                                   & \multicolumn{3}{c|}{\shortstack{Tested on\\$\text{\GPD}^{\text{DeDe}}+ \text{\GPD}^{\text{DeEn}}$}}                                                     \\ 
\hline
${\text{En+De}}$ & -4\%                    & -10.4\%                    & -4.9\%                    & -7.5\%                    & -11.9\%                    & -7\%                     \\ 
\hline
${\text{EnDe+DeEn}}$ & -3.3\%                    & -8.5\%                    & -4.4\%                    & -5.1\%                    & -9.8\%                    & -5.6\%                     \\ 
\hline
En+EnDe+ De+DeEn & -2.3\%                    & -8.5\%                    & -4.1\%                    & -2.8\%                    & -9.1\%                    & -4.2\%                     \\
\hline
\end{tabular}}
\caption{Mixed-language {\ASS} Performance}
\label{exp:mixedlanguage}
\vspace{-1em}
\end{table}

\subsection{Human Evaluation}

We conduct a human evaluation on a set of 1,827 production questions in German with candidate answers retrieved from relevant German documents.
We use the model fine-tuned on $\text{\GPD}^{\text{En+De+Fr+Es+It}}$.
The results show a marginal loss of only 3.07\% with respect to a multilingual QA system that uses the English best model to select German candidates translated in English.

\section{Conclusion}

We have presented a study on {\ASS} for multilingual QA systems.
First and foremost, we described our approach for creating a large-scale dataset for QA of a total 120K question-answer pairs for English.
We then presented a multilingual Transformer solution for {\ASS} consisting in (i) automatically transferring training data to different languages using MT; and (ii) applying a novel fine-tuning strategy using data translated in all languages along with mix language pair data.
We show that this approach enables the use of just one model with minimum performance drop
2.8\%
for German language, while the standard approach of using multilingual or monolingual fine-tuning would have had a drop of 10\%.
The use of mix-pair fine-tuning opens promising future research directions and applications for multilingual QA.

\bibliography{mqa}
\bibliographystyle{acl_natbib}

\end{document}